\documentclass{ifacconf}

\usepackage{graphicx}      
\usepackage{natbib}        
\usepackage{comment}
\usepackage{amsfonts}
\usepackage{amsmath}
\usepackage{amssymb}
\usepackage{array}
\usepackage{cancel}
\usepackage{color}
\usepackage{epsfig}

\usepackage{tikz} 
\usepackage{pgfplots}
\pgfplotsset{compat=1.17}

\begin{document}
\begin{frontmatter}

\title{Stabilization of vertical motion of a vehicle on bumpy terrain using deep reinforcement learning\thanksref{footnoteinfo}} 

\thanks[footnoteinfo]{This work was supported by the Automotive Research Center (ARC), a US Army Center of Excellence for modeling and simulation of ground vehicles, under Cooperative Agreement W56HZV-19-2-0001 with the US Army DEVCOM Ground Vehicle Systems Center (GVSC).}

\thanks[opsec]{DISTRIBUTION A. Approved for public release; distribution unlimited.
OPSEC \#: 6308}

\author[First]{Ameya Salvi} 
\author[Second]{John Coleman}
\author[Second]{Jake Buzhardt} 
\author[First]{Venkat Krovi}
\author[Second]{Phanindra Tallapragada}

\address[First]{Department of Automotive Engineering,}
\address[Second]{Department of Mechanical Engineering, 
   Clemson University, SC, 29634, USA (e-mail: ptallap@clemson.edu)}

\begin{abstract}                
Stabilizing vertical dynamics for on-road and off-road vehicles is an important research area that has been looked at mostly from the point of view of ride comfort. The advent of autonomous vehicles now shifts the focus more towards developing stabilizing techniques from the point of view of onboard proprioceptive and exteroceptive sensors whose real-time measurements influence the performance of an autonomous vehicle. The current solutions to this problem of managing the vertical oscillations usually limit themselves to the realm of active suspension systems without much consideration to modulating the vehicle velocity, which plays an important role by the virtue of the fact that vertical and longitudinal dynamics of a ground vehicle are coupled. The task of stabilizing vertical oscillations for military ground vehicles becomes even more challenging due lack of structured environments, like city roads or highways, in off-road scenarios. Moreover, changes in structural parameters of the vehicle, such as mass (due to changes in vehicle loading), suspension stiffness and damping values can have significant effect on the controller's performance. This demands the need for developing deep learning based control policies, that can take into account an extremely large number of input features and approximate a near optimal control action. In this work, these problems are addressed by training a deep reinforcement learning agent to minimize the vertical acceleration of a scaled vehicle travelling over bumps by controlling its velocity. 

\end{abstract}


\end{frontmatter}

\section{Introduction}
\begin{figure}[t]
\begin{center}
\includegraphics[width=8.4cm]{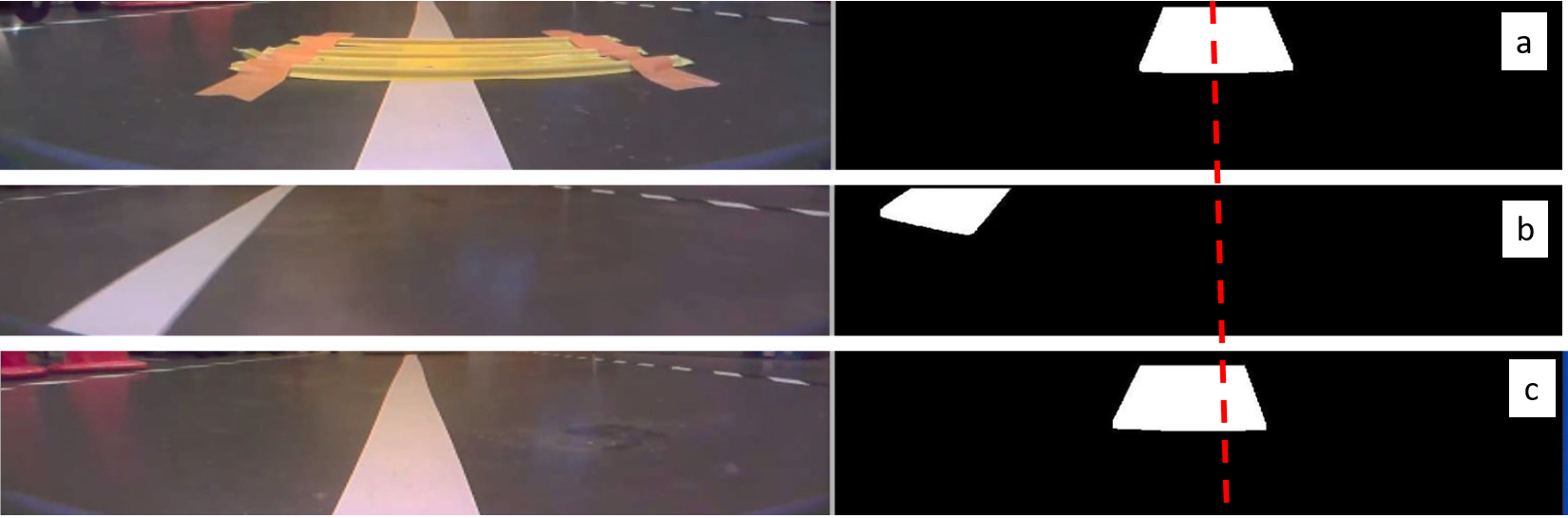}    
\caption{(a) A scaled vehicle running a vision based lane centering algorithm and approaching a bump.(b) The vehicle deviating from the path after hitting the bump at 5 m/s.(c) The vehicle deviating from the path after hitting the bump at 0.1 m/s.} 
\label{fig:BumpDisturbanceLC}
\end{center}
\end{figure}

Significant vertical oscillations are induced in a vehicle when it drives over an uneven terrain. If kept unchecked, these vertical oscillations can have an impact ranging from passenger discomfort to structural damage to delicate vehicular electronics [\cite{Elbanhawi2015}]. 
Remarkable disturbance is also generated for onboard sensors resulting in noisy or false measurements and impeding the overall performance of advanced driver assist technologies or related safety features. 
This problem now intensifies for autonomous vehicles which rely predominantly on perception information like camera images and lidar scans, resulting in sub-optimal or even unsafe driving conditions. 
This is illustrated in Fig.~\ref{fig:BumpDisturbanceLC}, which is a compilation of screen captures of a scaled vehicle running a vision based lane-centering algorithm and going over bumps. 
The figure illustrates how at higher velocity \ref{fig:BumpDisturbanceLC}(b), the tracking algorithm has to deal with significantly higher disturbance as compared to lower velocities \ref{fig:BumpDisturbanceLC}(c) when traversing a same bump. This simple example motivates the use of the longitudinal velocity of the vehicle as a control input in a framework with the objective of reducing the vertical oscillations of the vehicle chassis while traversing an uneven terrain. 

The configuration of a ground vehicle couples the longitudinal and vertical dynamics of the system making the solution to the stabilization problem both a function of vehicle speed and suspension parameters [\cite{milliken96:_race_car_vehic_dynam}]. 
It is conventional in the automotive industry to solve this problem using suspension tuning techniques at the vehicle design stage. The suspension stiffness and damping values are tuned with a significant safety factor for passenger comfort taking into account acceptable driving speeds and expected road bumps.
It has been concluded that limiting oscillations of the  vehicle's vertical acceleration to a frequency range of 0-65 rad/s ensures passenger comfort [\cite{Ian2000}]. Researchers have also tried to implement advanced control techniques varying from structurally changing the suspension configuration [\cite{Ryazantsev_2020}] to incorporating active or semi-active suspension systems [\cite{Zhi2021}] which modify the suspension characteristics in real-time.
Combination controllers have also been investigated such as combining a PID controller (which focuses on minimizing the vertical acceleration) and a Fuzzy logic controller(which focuses on minimizing pitch) [\cite{Yu20001}]. Preview-based suspension control is the latest research trend in this field where the community is now trying to make adaptive suspensions based on terrain preview [\cite{THEUNISSEN2021206}]. 

Unfortunately, all of these techniques heavily rely on adopting complex electronic suspension systems which are not readily available in all vehicles. In absence of such systems vehicle longitudinal velocity becomes the only actuation parameter that can mitigate the vertical oscillations over a rough terrain. On-road autonomous vehicles may benefit to some extent from predefined traffic regulations that specify travel speed over speed bumps or rumbling strips but velocity control of off-road ground vehicles for stabilizing vertical oscillations still remains an important problem.

Off-road ground vehicles face significant wear over their life cycle. Structural parameters like mass, suspension characteristics and tire performance will change over time [\cite{Farroni}]. Many of these changes are difficult to measure or quantify in real-time but can impact the overall dynamics of the vehicle. In such a scenario, it becomes relevant to have adaptive control techniques that can achieve optimal or near-optimal control performance by adapting or learning solely from the observable data. Deep reinforcement learning has shown tremendous potential when it comes to optimal, adaptive and learning-based control scheme [\cite{Sutton1991}]. 
In absence of an accurate physics-based, ever-changing system parameters, deep reinforcement learning can leverage its model free strategy for learning approximate optimal solutions [\cite{Degris}].

A deep reinforcement learning based control policy that tries to minimize the vertical oscillations for the vehicle going over bumps has been demonstrated in this paper using a one tenth scaled vehicle. 
The rest of the paper is organized as follows: the first section discusses developing a simplified half-car model for the actual scaled vehicle and setting it up for simulation using a deep reinforcement learning framework. Next, dynamic reward shaping is discussed and the simulation results are studied. The final section elaborates on the actual deployment of the policy on the scaled vehicle platform and validating the results.

\section{Problem formulation}

\begin{figure}
\begin{center}
\includegraphics[width=0.8\linewidth]{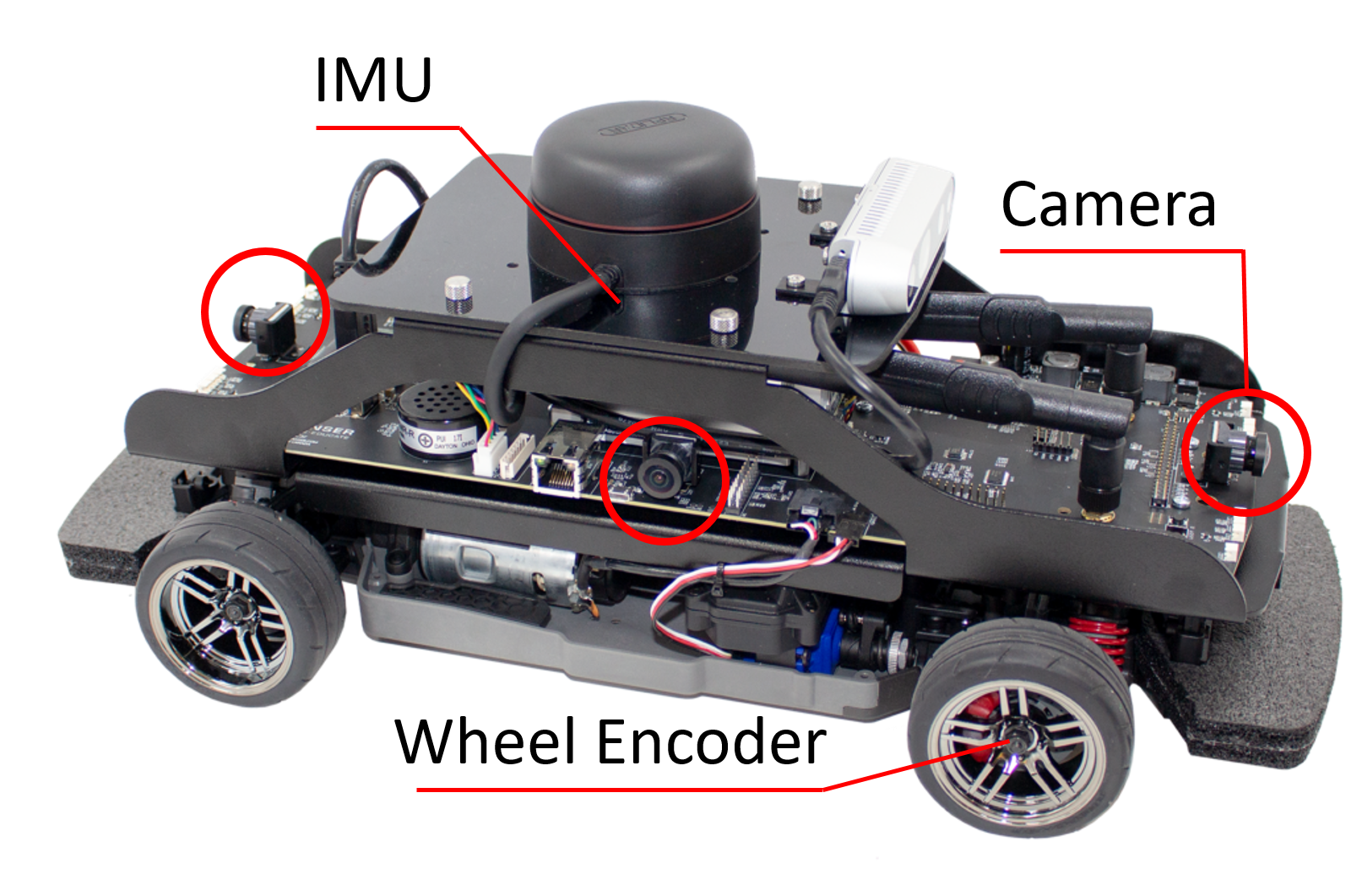}    
\caption{Quanser QCar: Scaled Autonomous Vehicle} 
\label{fig:QCar}
\end{center}
\end{figure}

\subsection{Problem set up}
\begin{figure}
\begin{center}
\includegraphics[width=8.4cm]{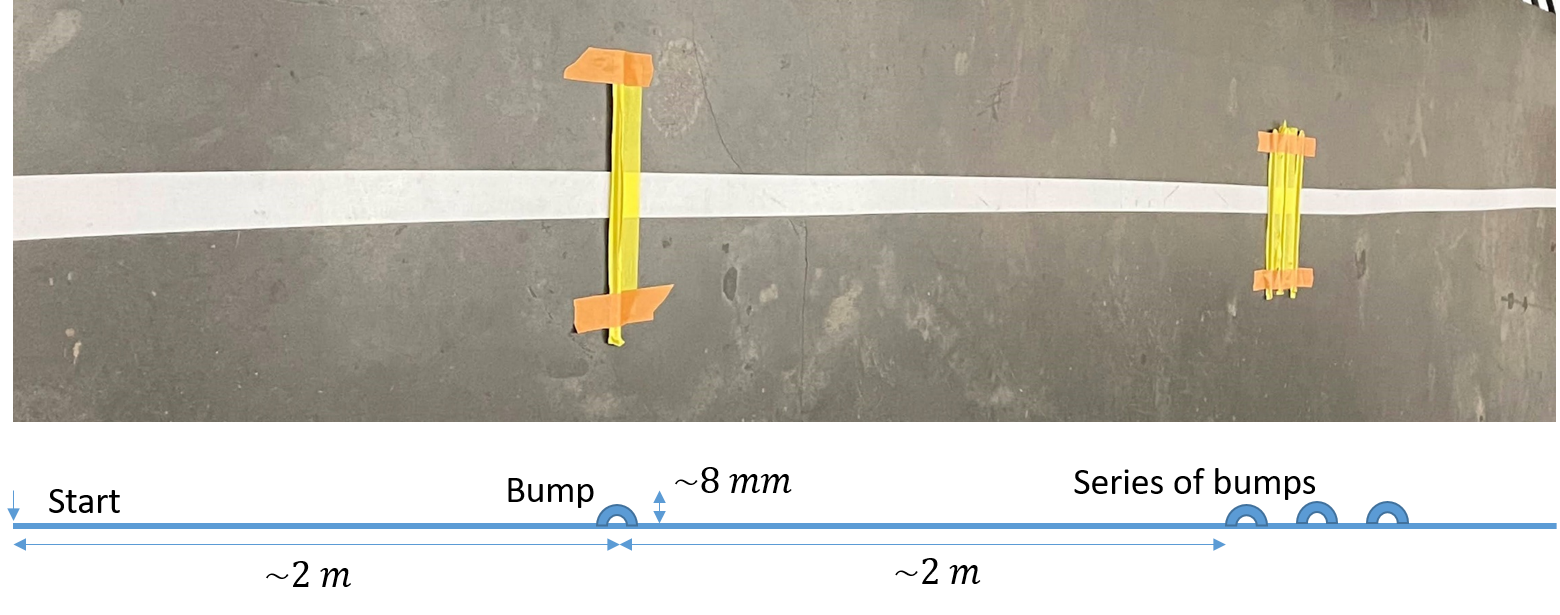}    
\caption{Track representation} 
\label{fig:TrackSchematic}
\end{center}
\end{figure}

\begin{figure}
\begin{center}
\includegraphics[width=0.9\linewidth]{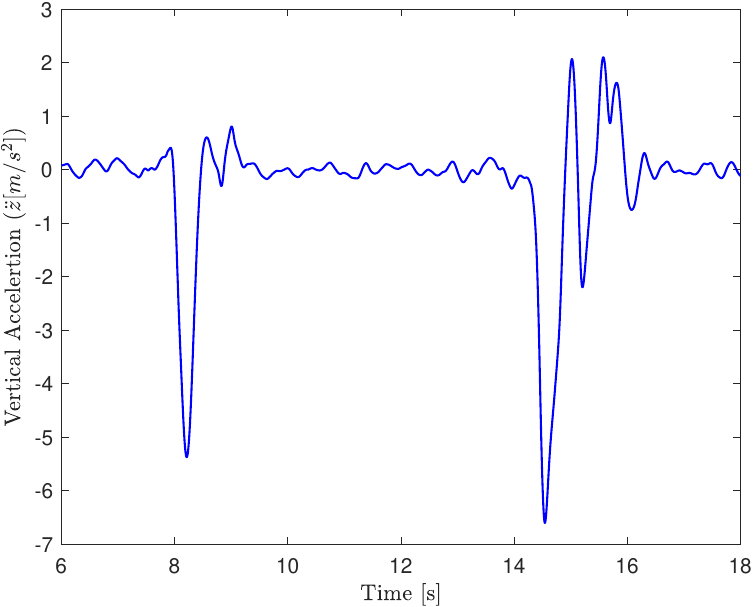}    
\caption{Vertical acceleration readings when traversing the track while tracking a constant longitudinal velocity of 1 m/s} 
\label{fig:VertAccOL}
\end{center}
\end{figure}

As shown in Fig.~\ref{fig:BumpDisturbanceLC}, vehicle stabilization tasks are highly dependent on the vehicle velocity. 
In many cases, longitudinal vehicle velocity is the only available control parameter for stabilizing the vertical dynamics of the vehicle. Hence, in this work the primary problem is to minimize the vertical accelerations of a scaled vehicle going over bumps by controlling its longitudinal velocity based on terrain preview information provided by an onboard camera. A scaled vehicle platform, Quanser QCar, is used as a test bench for developing the reinforcement learning based control policies. The vehicle tries to traverse a straight path with irregularly placed bumps as shown in Fig.~\ref{fig:TrackSchematic} with the objective of tracking a specified constant velocity and minimizing the vertical accelerations when going over the bumps. The scaled vehicle is equipped with cameras for bump preview, an inertial measurement unit (IMU) for getting the vertical accelerations, and wheel encoders for measuring the realized longitudinal velocity. Figure \ref{fig:VertAccOL} represents the deviations from nominal vertical acceleration (9.8 m/s$^2$) when the scaled vehicle traverses the track represented in Fig.~\ref{fig:TrackSchematic} while tracking a constant longitudinal velocity of 1 m/s.
\begin{table}[h!]
\caption{Model Parameters}
\begin{center}
\label{tab:Model_Parameters}
\begin{tabular}{cc}
Parameter & Value \\\hline
Mass of QCar & 1.391 kg \\[1ex]
$L_1$, $L_2$ & 0.128 m\\[1ex]
Damping of QCar & 77.6 N\,s/m \\[1ex]
Stiffness of QCar & 19.6 N/m\\[1ex]
Moment of Inertia of QCar & 0.001897 $\mathrm{kg} \cdot \mathrm{m}^{2}$\\[1ex]
Max Bump Height & 0.008 m\\\hline
\end{tabular}
\end{center}
\end{table}

\subsection{Surrogate Model}

\subsubsection{Half-car model:}
\begin{figure}
\begin{center}
\includegraphics[width=8.4cm]{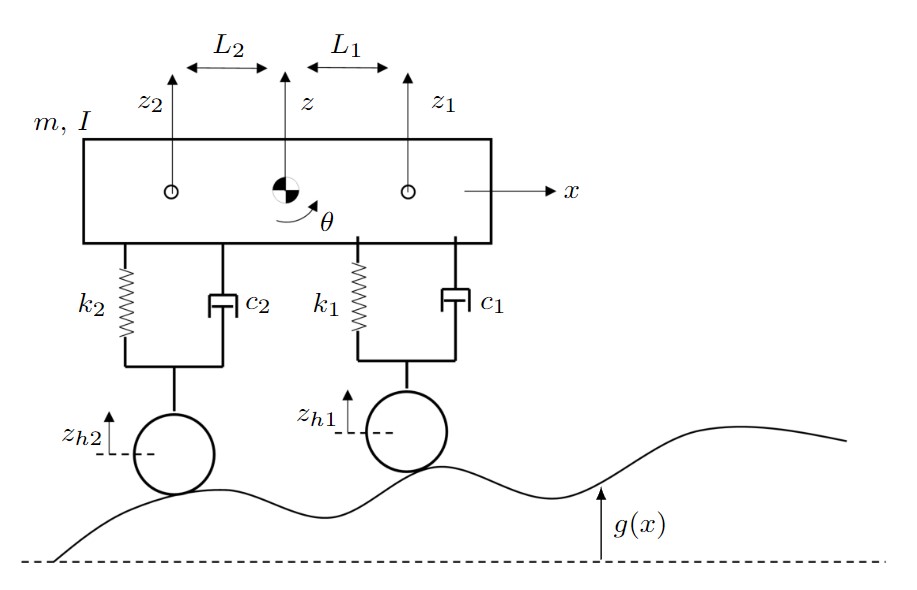}    
\caption{Half-car vehicle model} 
\label{fig:half car}
\end{center}
\end{figure}

A half-car model with passive suspensions was used to model the QCar's vertical and pitching dynamics, as shown in Fig.~\ref{fig:half car}.  We use the subscripts 1 and 2 here to denote the front and rear axle of the vehicle, respectively.
The equations of motion for the half-car are derived by summing forces and moments on the vehicle body.  The coordinates considered are the vertical displacement $z$ and pitch angle $\theta$ of the vehicle, both describing the motion of the center of mass of the vehicle. 
These equations are given as follows.
\begin{align} 
\label{eq:Angular Acceleration}
I\ddot{\theta} &= - k_{1}(z_{1}-z_{h1})L_{1} + k_{2}(z_{2}-z_{h2})L_{2} \\
&\qquad - c_{1}(\dot{z}_{1}-\dot{z}_{h1})L_{1} + c_{2}(\dot{z}_{2}-\dot{z}_{h2})L_{2} 
\nonumber\\[1ex]
\label{eq:Vertical Acceleration}
m\ddot{z} &= -k_{1}(z_{1}-z_{h1}) - k_{2}(z_{2}-z_{h2}) \\
& \qquad - c_{1}(\dot{z}_{1}-\dot{z}_{h1}) - c_{2}(\dot{z}_{2}-\dot{z}_{h2})  \nonumber
\end{align}
Here the linear stiffnesses of the front and rear suspensions are given by $k_{1}$ and $k_{2}$ and damping coefficients by $c_{1}$ and $c_{2}$. The total mass of the car and moment of inertia are denoted by $m$ and $I$ respectively. The wheelbase of the car is broken down into $L_{1}$ and $L_{2}$, with $L_{1}$ being the distance from the center of gravity to the front wheel, and $L_{2}$ the distance from the center of gravity to the rear wheel. 
The vertical displacements of the front and rear wheels from equilibrium due to the terrain are $z_{h1}$ and $z_{h2}$ respectively. 
The vertical displacements of the chassis from equilibrium at the front and rear axles are $z_{1}$ and $z_{2}$, respectively, which can be written in terms of $z$ and $\theta$ as follows. 
\begin{equation*} \label{eq:z}
\begin{array}{ll}
z_{1}=z-L_{1}\cos{\theta} \\
z_{2}=z+L_{2}\cos{\theta}.
\end{array}
\end{equation*}
We model the terrain profile here as a continuous function of the longitudinal displacement of the vehicle, $x$.  So, the displacement of each wheel from equilibrium is given by
$z_{hi}=g(x_i)$ for $i\in\{1,2\}$.
Then by chain rule, the corresponding wheel velocity is written as $\dot{z}_{hi} = g^\prime(x_i)\dot{x}_i$, where the longitudinal position of the front and rear axles are given by $x_1 = x + L_1\cos\theta$ and $x_2 = x - L_2\cos\theta$.

In order to model a terrain composed of discrete bumps, as shown in Fig.~\ref{fig:TrackSchematic}, we take the function $g(x)$ 
to be a summation of Gaussian functions, with each Gaussian representing the profile of a bump. That is, for a terrain composed of a series of $N_b$ bumps centered around the mean positions $\mu_j$ and with spread characterized by variances $\sigma_j^2$, we would have
\begin{equation*} \label{eq:zh}
g(x) = \sum_{j=1}^{N_b}H_j \exp\left(\frac{-(x-\mu_j)^2}{2\sigma_j^2}  \right) .
\end{equation*}

Finally, we take the longitudinal dynamics to be described by a first-order delay on a commanded velocity, $u_x$.  That is, 
\begin{equation}\label{eq:Command Velocity}
    \tau \ddot{x} + \dot{x} = u_x
\end{equation}
with $\tau$ being the time constant. 
In the reinforcement learning framework, we take the command velocity, $u_x$ to be the action chosen by the agent.

The numerical values used in simulation for the constant parameters in this model are given in Table \ref{tab:Model_Parameters}.


\subsection{Deep Deterministic Policy Gradient}\label{sec:ddpg}
\begin{figure}
\begin{center}
\includegraphics[width=8.4cm]{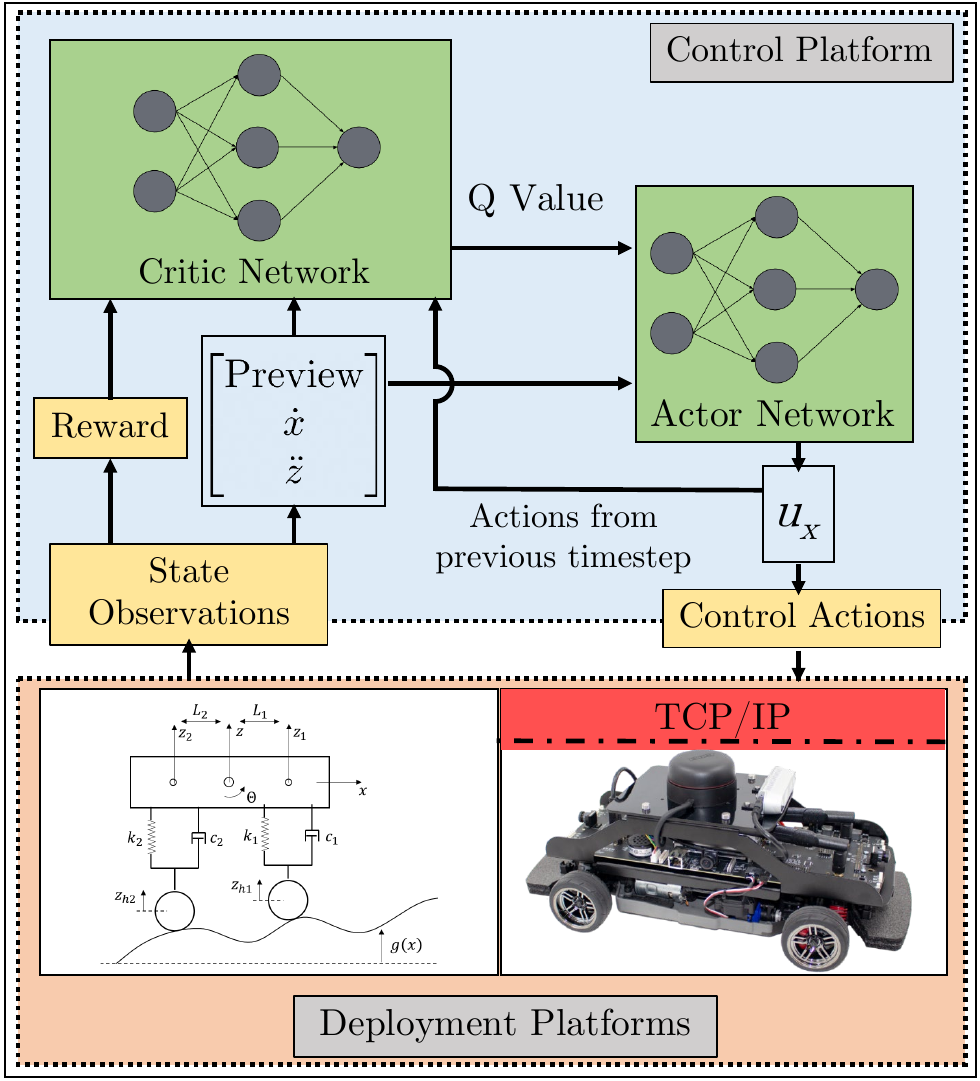} 
\caption{Modular representation of the deep deterministic policy gradient (actor-critic network) interacting with the simulation and scaled vehicle platforms} 
\label{fig:RLOverview}
\end{center}
\end{figure}

The deep deterministic policy gradient (DDPG) [\cite{Lillicrap15_DDPG}] is a reinforcement learning agent that can deal with continuous state and action spaces. 
As shown in Fig.~\ref{fig:RLOverview}, the agent is comprised of two deep neural networks, actor and critic, which work in tandem during the training process. 
The objective of the training process is to learn a control policy such that the vehicle tracks a prespecified desired velocity, $\dot{x}_{d}$, while also minimizing the vertical acceleration, $\ddot{z}$. 
In the current framework, the vehicle velocity, $\dot{x}$, the vertical acceleration, $\ddot{z}$, and terrain preview are provided as state observations and the command velocity, $u_x$ is the control action. 
The terrain preview is provided to the agent as a unitless quantity that is indicative of the approaching bumps. 
In practice, this is calculated as the proportion of the image in which the bump is seen.  
This number is inversely proportional to the distance of the vehicle from the bump, as when the vehicle is closer to the bump, the bump occupies a larger share of the camera view.  
In the experimental setup, 
the bumps are covered by a yellow tape 
and a simple threshold is applied to the HSV image to obtain a binary image identifying the bump in the camera view. 
The terrain preview quantity is then calculated as the ratio of number 
of pixels identified as the bump by the threshold to the total number of pixels in the image.
Figure \ref{fig:Preview} shows the preview values received by the agent in a single training episode.
The peaks in the plot are an indicator of number of upcoming bumps. 

Once the training is complete, the actor network itself is deployed as the control policy which predicts control action based on provided observations. 
The MATLAB-Simulink platform is used for setting up the training process.
This provides the flexibility to interface with the simulation frameworks and the actual hardware. As discussed in earlier sections, the majority of the training first takes place in the simulation framework using the simplified half-car model as a learning environment. 
As referenced in Fig. \ref{fig:RLOverview}, the same tool-chain can now also interact with the QCar over a TCP/IP connection for further training or deployment.

\subsubsection{Reward function:}
The notion of reward function is similar to a cost function, where there is a parameterized objective function, which influences the efficient and correct convergence of the learning algorithm. The DDPG agent learns a policy such that this reward function is maximized. 
Eq. \ref{eq:Reward} gives a nominal form of the reward function employed here.  This form is chosen so that the maximum value will be zero when the vertical acceleration, $\ddot{z}$, is equal to 9.8 m/s$^2$ and the longitudinal velocity, $\dot{x}$, is equal to the desired velocity, $\dot{x}_{d}$. 
\begin{equation} \label{eq:Reward}
\begin{array}{ll}
R=-w_{1}(\ddot{z}-9.8)^{2}-w_{2}(\dot{x}-\dot{x}_{d})^{2}
\end{array}
\end{equation}
Here $w_1$ and $w_2$ are weights that quantify the relative importance of the two terms in the reward.
Reward shaping is a popular technique in which human bias from access to certain information can be explicitly modelled in the reward function to improve the learning process. 
In this work, a dynamic reward shaping approach is implemented which modulates the weights, $w_{1}$ and $w_{2}$, of the reward function parameters based on the preview information. 

\begin{figure}
\begin{center}
\includegraphics[width=0.9\linewidth]{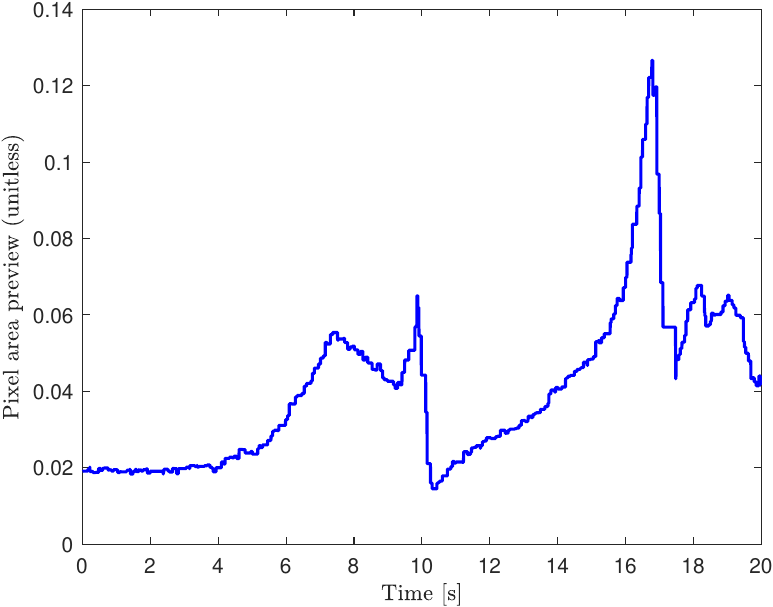}    
\caption{Preview of the bump as associated with the binary pixel information} 
\label{fig:Preview}
\end{center}
\end{figure}

Three variations of the reward function, given by Eq. \ref{eq:RewardA} to Eq. \ref{eq:RewardC} were studied for minimizing the peak and RMSE acceleration values while tracking a certain velocity. 
In the statically weighted reward given by Eq. \ref{eq:RewardA}, the weights for the function parameters stay constant throughout the simulation episode.
\begin{equation} \label{eq:RewardA}
\begin{array}{ll}
R=-(\ddot{z}-9.8)^{2}-75(\dot{x}-\dot{x}_{d})^{2}
\end{array}
\end{equation}
In the conditionally weighted reward given by Eq. \ref{eq:RewardB}, the weights change if a certain condition is met.  This condition is based on the pixel area fraction discussed in Sec. \ref{sec:ddpg}, shown in Fig. \ref{fig:Preview}, and denoted here by $p$. 
\begin{equation} \label{eq:RewardB}
    R = \begin{cases}   
    100(\ddot{z}-9.8)^{2}-75(\dot{x}-\dot{x}_{d})^{2} \qquad & \mathrm{if}\, p >0.05\\[1ex]
    -(\ddot{z}-9.8)^{2}-75(\dot{x}-\dot{x}_{d})^{2} \qquad & \text{otherwise}
    \end{cases}
\end{equation}
In the function weighted reward, Eq. \ref{eq:RewardC}, the weights are a function of the bump preview and vary linearly with the pixel preview quantity.
\begin{equation} \label{eq:RewardC}
\begin{split}
R&=-w(p)(\ddot{z}-9.8)^{2}-75(\dot{x}-\dot{x}_{d})^{2}\\
w(p) &= 100\,p
\end{split}
\end{equation}
In the latter two reward formulations, the vertical acceleration is penalized more as the vehicle approaches the bump. The agent is trained for a total of 500 episodes in the simulation framework with training parameters mentioned in table \ref{tab:Training_Parameters}.
\begin{table}[h!]
\caption{Training Parameters}
\begin{center}
\label{tab:Training_Parameters}
\begin{tabular}{cc}
Parameter & Value \\\hline
Noise variance & 0.8 \\[1ex]
Noise decay rate & $1e^{-4}$ m\\[1ex]
Policy learning rate & $1e^{-4}$ \\[1ex]
Smoothing Factor & $1e^{-3}$ m\\\hline
\end{tabular}
\end{center}
\end{table}

%

\subsection{Sim2Real Deployment}\label{Sim2Real_Deployment}
Since the simulation environment is idealistic and prone to modelling inaccuracies, the agent is supplemented by additional training on the physical vehicle. 
This is done so that the idealistic policy learned in the simulation will adapt to actual vehicle parameters, sensor information and communication delays making it an adaptive learning-based control. 
Hence, after simulator training, the agent was further trained for 
an additional
20 episodes using the scaled vehicle. 
MATLAB - Simulink code generation capability was used for deployment on the QCar hardware.
During deployment, the QCar communicated wirelessly with a host machine providing actuation signals (command velocity) based on the observation data received from the Qcar (IMU and image pixel information). The communication frequncy was maintained at 120 hz.

\section{Results and Discussion}
\begin{figure}
\begin{center}
\includegraphics[width=0.9\linewidth]{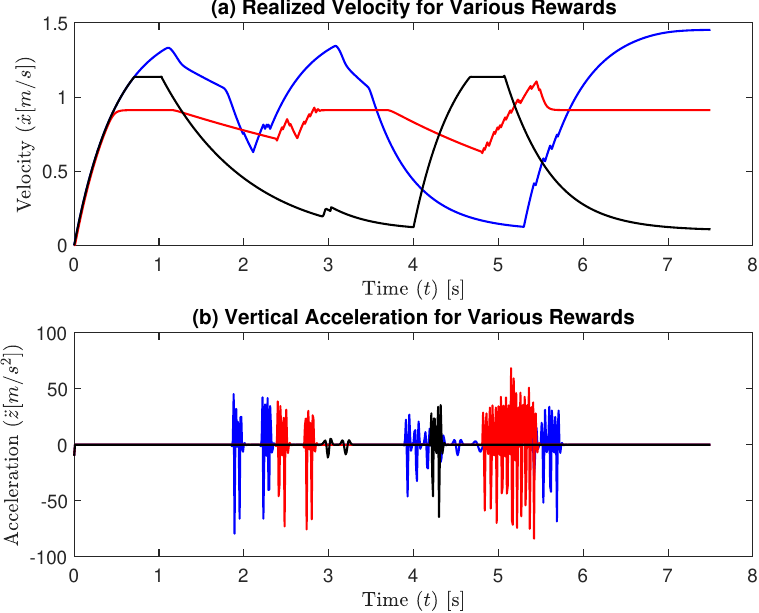}    
\caption{Simulation Results : (a) Realized Velocity (b) Vertical Acceleration (Blue : statically weighted reward (Eq. \ref{eq:RewardA}), Red: Condition weighted reward (Eq. \ref{eq:RewardB}), Black: Function weighted reward (Eq. \ref{eq:RewardC}))} 
\label{fig:SimulationResultsRewardShaping}
\end{center}
\end{figure}
\subsection{Simulation Results}
Figure \ref{fig:SimulationResultsRewardShaping} shows a performance comparison of the policies trained with the three different reward formulations (Eqs. \ref{eq:RewardA}-\ref{eq:RewardC}) applied in the simulation environment. When trained for exactly same duration, the agent trained with function weighted reward develops a policy that penalizes the vertical acceleration more and as a result shows significant reduction in the peak as well as root mean squared error values for acceleration without much drop in the velocity tracking performance as compared to the other two. 
For this reason, the agent with function weighted reward was selected for further training and deployment on the scaled vehicle.

\subsection{Sim2Real Adaptation}
\begin{figure}
\begin{center}
\includegraphics[width=0.9\linewidth]{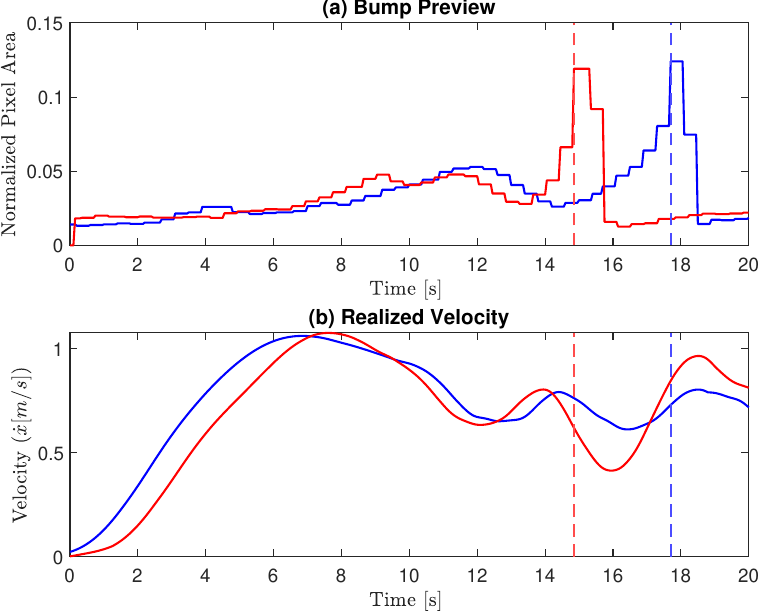}    
\caption{Agent adapting to real data: Directly after simulation (Blue) and after 20 training episodes (Red)} 
\label{fig:AdaptTrain}
\end{center}
\end{figure}

With the hardware configuration mentioned in section \ref{Sim2Real_Deployment}, the scaled vehicle platform is trained for 20 more episodes on physical bumps. Figure \ref{fig:AdaptTrain} shows the policy improvement over the episodes.
The blue line shows the policy directly out of simulation and the red line indicates after 20 additional training episodes on the scaled vehicle. 
After this additional training, the agent learns to command much lower velocity over the series of bumps as compared to the simulation-only case. 
This can be seen by observing the blue and red vertical dashed lines in Fig. \ref{fig:AdaptTrain}, which indicate where the series of bumps occurs in each case.
Figure \ref{fig:AdaptTrain}(a) shows the pixel ratio preview of the bump as seen by the agent. The actual bump occurs slightly after the preview signal peaks. 
It can be observed that after 20 episodes of training on the scaled vehicle, the vehicle slows down slightly after the bump preview whereas without this training it slows down just as the bump is in sight.   

\subsection{Results}
\begin{figure}
\begin{center}
\includegraphics[width=0.9\linewidth]{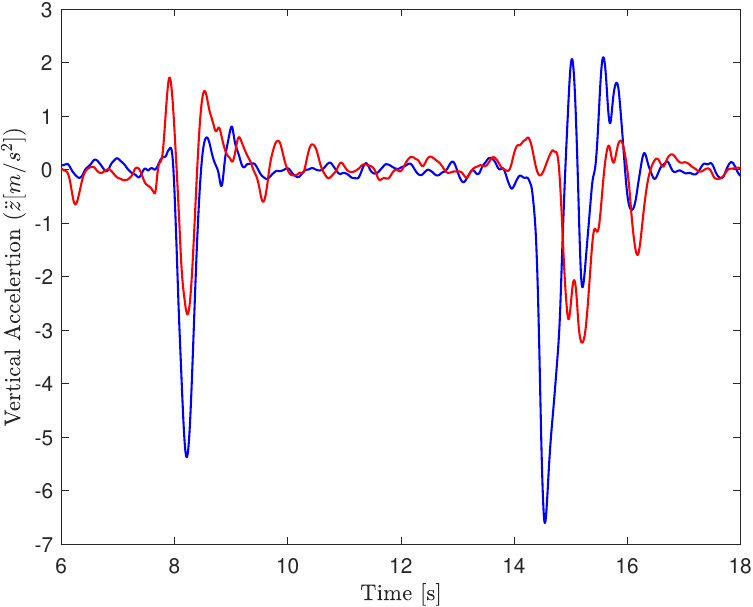}    
\caption{Comparison of Acceleration values : Directly after simulation (Blue) and after 20 training episodes (Red)}
\label{fig:VAOLCLCompare}
\end{center}
\end{figure}

The plots in Fig. \ref{fig:VAOLCLCompare} show the vertical acceleration readings for the vehicle going over the bumpy track. The blue dotted line are from an open loop case where the vehicle is given a constant command velocity of 1 m/s without any influence of the control policy. The solid red line represents the values when the vehicle travels at the speed modulated by the reinforcement learning policy. 
The peaks are slightly offset as in the latter case the policy commands a lower velocity thus delaying the encounter with the bumps. 
It can be observed that the peak acceleration values are significantly reduced when the velocity is controlled by the reinforcement learning controller.

\subsection{Validation}
\begin{figure}
\begin{center}
\includegraphics[width=0.9\linewidth]{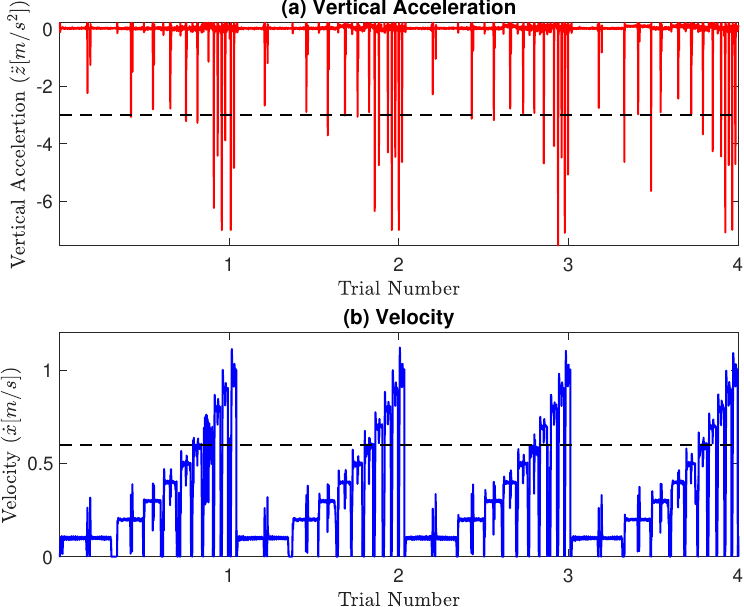}    
\caption{Peak acceleration values (Red) at various velocities (Blue) when going over bumps without velocity control} 
\label{fig:ExperimentTrials}
\end{center}
\end{figure}

The velocity profile realized by the policy does seem to reduce the vertical accelerations as compared to the constant velocity case but it can always be argued that going as slow as possible would be the solution to minimizing the vertical acceleration. To verify if this is really the case, the vehicle was made to go over a bump for velocities ranging from 0.1 m/s to 1 m/s at increments of 0.1 m/s. This was conducted for multiple trials and the data is plotted in Fig. \ref{fig:ExperimentTrials}. 
Figure \ref{fig:ExperimentTrials}(b) shows the velocities used for trial and  \ref{fig:ExperimentTrials}(a) shows the resultant vertical acceleration when going over the bump. It can be observed from this that at velocities lower than roughly 0.6 m/s (indicated by the dashed line in Fig. \ref{fig:ExperimentTrials}(b)) further velocity reduction 
doesn't have a significant effect on the vertical acceleration values. 
The dashed line in Fig. \ref{fig:ExperimentTrials}(a) shows that roughly -3 m/s$^2$ is the maximum possible drop in peak vertical acceleration values that can be obtained by reducing the forward velocity. This information helps conclude that ideally, the controller must command velocity such that the realized velocity should be in the range of 0.6 m/s to 1.0 m/s. Going anything below 0.6 m/s won't have a significant impact on the acceleration values. It can be observed from the figure \ref{fig:AdaptTrain}(b) that the realized velocity stays in a similar range of 0.5 m/s - 1 m/s, thus indicating that the control policy learned is non-trivial.


\section{Conclusion and future work}
In this work, a DDPG agent is trained to develop a control policy for minimizing vertical acceleration for a scaled vehicle going over bumps. The primary motivation for the work is to develop solutions for conventional vehicles or mobile robot platforms that do not have access to active suspension systems or articulated wheel structures. 
As a future step, it could be beneficial to develop an end-to-end framework where instead of binary pixel information for the bump, a raw image data can be used as a feature input. A future work might include using the raw image and the binary image both as a feature input to study if it makes the policy any more robust to photo-sensitivity. 
Additionally, future lines of research could consider how this algorithm could couple with more advanced autonomy and motion planning, such as obstacle detection and avoidance or tracking of more complex paths. As an extension to full scaled vehicles, certain aspects of ergonomics and passenger comfort can be incorporated as constraints in the reward function to avoid jerks while providing command velocity. 

\begin{ack}
Clemson University, Department of Mechanical and Automotive engineering acknowledge the technical and financial support of the Automotive Research Center (ARC) in accordance with Cooperative Agreement W56HZV-19-2-0001 US Army CCDC Ground Vehicle Systems Center (GVSC) Warren, MI.
\end{ack}

\bibliography{MECC_2022_Ref}             
                                                   







\end{document}